\documentclass[10pt,twocolumn,letterpaper]{article}

\usepackage[pagenumbers]{cvpr} 

\usepackage{graphicx}
\usepackage{amsmath}
\usepackage{amssymb}
\usepackage{booktabs}
\usepackage{enumitem}
\usepackage[numbers]{natbib}

\usepackage[table]{xcolor}
\definecolor{lightgray}{rgb}{0.85, 0.85, 0.85}
\newcolumntype{g}{>{\columncolor{lightgray}}c}

\usepackage[pagebackref,breaklinks,colorlinks]{hyperref}

\usepackage[capitalize]{cleveref}
\crefname{section}{Sec.}{Secs.}
\Crefname{section}{Section}{Sections}
\Crefname{table}{Table}{Tables}
\crefname{table}{Tab.}{Tabs.}

\newcommand{\x}{\mathbf{x}}
\newcommand{\y}{\mathbf{y}}
\newcommand{\z}{\mathbf{z}}
\newcommand{\M}{\mathbf{M}}
\newcommand{\W}{\mathbf{W}}
\newcommand{\A}{\mathbf{A}}
\newcommand{\LL}{\mathbf{L}}

\newcommand{\R}{\mathbb{R}}

\newcommand{\up}[1]{\scriptsize(+#1)}
\newcommand{\down}[1]{\scriptsize(-#1)}
\newcommand{\bup}[1]{{\color{blue}\scriptsize(+#1)}}
\newcommand{\rdown}[1]{{\color{red}\scriptsize(-#1)}}
\newcommand{\buparrow}{{\color{blue}$\pmb{\uparrow}$}}
\newcommand{\rdownarrow}{{\color{red}$\pmb{\downarrow}$}}

\newcommand{\blank}{\phantom{\scriptsize(+x.xx)}}

\def\cvprPaperID{*****} 
\def\confName{CVPR Transformers for Vision Workshop}
\def\confYear{2022}

\begin{document}

\title{OAMixer: Object-aware Mixing Layer for Vision Transformers}

\author{
Hyunwoo Kang\thanks{Equal contribution}$\:\:$ \quad Sangwoo Mo$^*$ \quad Jinwoo Shin \\
KAIST\\
{\small \texttt{\{hyunwookang,swmo,jinwoos\}@kaist.ac.kr}} \\
}
\maketitle

\begin{abstract}
Patch-based models, e.g., Vision Transformers (ViTs) and Mixers, have shown impressive results on various visual recognition tasks, alternating classic convolutional networks. While the initial patch-based models (ViTs) treated all patches equally, recent studies reveal that incorporating inductive bias like spatiality benefits the representations. However, most prior works solely focused on the location of patches, overlooking the scene structure of images. Thus, we aim to further guide the interaction of patches using the object information. Specifically, we propose OAMixer (object-aware mixing layer), which calibrates the patch mixing layers of patch-based models based on the object labels. Here, we obtain the object labels in unsupervised or weakly-supervised manners, i.e., no additional human-annotating cost is necessary. Using the object labels, OAMixer computes a reweighting mask with a learnable scale parameter that intensifies the interaction of patches containing similar objects and applies the mask to the patch mixing layers. By learning an object-centric representation, we demonstrate that OAMixer improves the classification accuracy and background robustness of various patch-based models, including ViTs, MLP-Mixers, and ConvMixers. Moreover, we show that OAMixer enhances various downstream tasks, including large-scale classification, self-supervised learning, and multi-object recognition, verifying the generic applicability of OAMixer.\footnote{
Code: \url{https://github.com/alinlab/OAMixer}
}
\end{abstract}

\section{Introduction}
\label{sec:intro}

Patch-based models, i.e., models that process an input image as a sequence of visual patches, have arisen as a new paradigm of neural networks for visual data, regarded as a promising alternative to convolutional neural networks (CNNs) \cite{lecun1998gradient}. Remarkably, patch-based models have achieved state-of-the-art results on various computer vision tasks by favoring the scaling properties \cite{dosovitskiy2020image,zhai2021scaling,abnar2021exploring}. Moreover, patch-based models merit multiple advantages, including out-of-distribution generalization \cite{naseer2021intriguing,bai2021transformers,paul2021vision}, a natural extension to video domains \cite{bertasius2021space,arnab2021vivit}, integration with other domains like language or speech \cite{radford2021learning,akbari2021vatt}, and easily combined with state-of-the-art visual self-supervised learning \cite{he2021masked}.

The core concept of patch-based models is to update patch-wise representations by alternating the computation \textit{within} patches and \textit{among} patches, called channel mixing and patch (or token) mixing, respectively. To this end, the design of patch mixing layers is widely investigated. The pioneering work named Vision Transformer (ViT) \cite{dosovitskiy2020image} considered self-attention \cite{vaswani2017attention}, and other works considered various mixing layers such as feed-forward \cite{tolstikhin2021mlp,touvron2021resmlp,liu2021pay}, convolution \cite{trockman2022patches}, or even pooling operation \cite{yu2021metaformer}. Most standard patch-based models use self-attention or feed-forward mixing layers, which minimizes the prespecified inductive biases by employing all patches equally \cite{khan2021transformers}.

While this data-centric approach is effective in large-scale scenarios, recent works claim that incorporating inductive biases is still vital for patch-based models, especially when learned from limited data \cite{steiner2021train}. To this end, numerous works incorporated spatial inductive bias for patch-based models following the analogy of convolutions. They can be categorized into two groups: designing a patch mixing layer utilizing the location of patches \cite{d2021convit,dai2021coatnet,wu2021cvt,trockman2022patches,lian2021mlp} or building an architecture that combines convolutional or pooling layers with patch mixing layers \cite{liu2021swin,yuan2021tokens,wang2021pyramid,fan2021multiscale,heo2021rethinking,yuan2021incorporating,xiao2021early}. However, both approaches focus on sample-agnostic spatial inductive bias, which overlooks the sample-specific object structures.

\begin{figure*}[t]
\centering
\includegraphics[width=.9\textwidth]{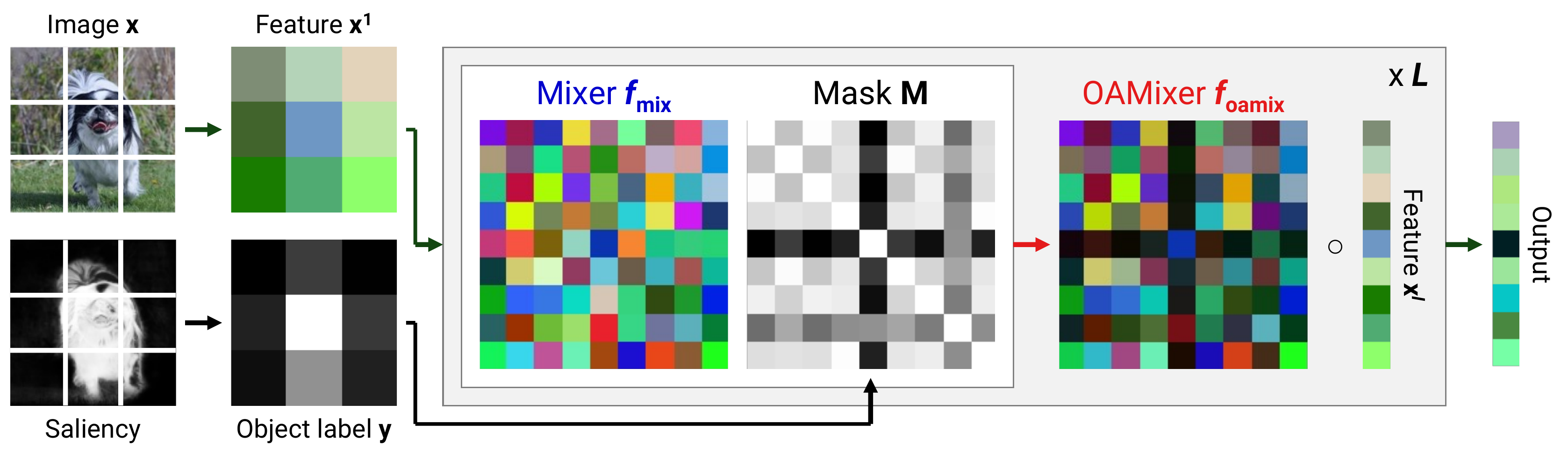}
\caption{
Conceptual illustration of our method, OAMixer. Given an image $\x$, we assume that patch-wise object labels $\y$ (either from pixel-level labels or given by patch-level directly) are available. Using this label, we compute a reweighting mask $\M$ which modifies the original patch mixing layer $f_\texttt{mix}$ to the data-specific \textit{object-aware mixed} layer $f_\texttt{oamix}$.
}\label{fig:concept}
\end{figure*}

\textbf{Contribution.}
We propose OAMixer, a novel reweighting scheme for patch mixing layers leveraging the object structure of images. The main idea of OAMixer is to strengthen the interaction of patches containing similar objects while regularizing the connection of different objects (and background). Intuitively, OAMixer improves the discriminability of objects (i.e., better classification) and reduces the spurious correlations between objects and backgrounds (i.e., robust to background shifts) by disentangling object representations. Figure~\ref{fig:concept} visualizes the high-level concept of OAMixer.

Our framework consists of two stages. First, we compute a reweighting mask for each image using the pairwise similarity of patches, where patch-wise object labels give the similarity. {Here, one can obtain the object labels in unsupervised or weakly-supervised manners, i.e., no additional human-annotating cost is necessary.\footnote{
One can also use a pretrained object labeler trained on some external dataset like ImageNet.
}} We then calibrate the interaction of patch mixing layers using the reweighting mask and a learnable scale parameter that controls the level of calibration. Although OAMixer can be applied to any patch-based model in principle, one needs a careful design to consider the interaction dynamics of each layer. Thus, we illustrate specific instantiations of OAMixer for three representative mixing layers: self-attention, feed-forward, and convolution.

In our experiments, we demonstrate the superiority of object-aware inductive bias from OAMixer:
\begin{itemize}
\item Improves classification and background robustness over various patch-based models, including ViTs \cite{dosovitskiy2020image}, MLP-Mixers \cite{tolstikhin2021mlp}, and ConvMixers \cite{trockman2022patches}.
\item Outperforms the patch mixing layers considering spatial inductive bias (e.g., ConViT \cite{d2021convit}).
\item Outperforms the prior methods utilizing patch-level supervision (e.g., TokenLabeling \cite{jiang2021all}).
\item Unlike prior methods (e.g., ConViT and TokenLabeling), OAMixer enhances the background robustness.
\end{itemize} 
OAMixer also benefits various downstream tasks:
\begin{itemize}
\item Improves large-scale classification, e.g., increases ImageNet \cite{deng2009imagenet} top-1 accuracy of DeiT-B \cite{touvron2021training} from 78.45\% to 82.18\% (+3.73\%).
\item Improves self-supervised learning, e.g., increases ImageNet linear probing of DeiT-T with DINO \cite{caron2021emerging} from 59.37\% to 61.16\% (+1.79\%).
\end{itemize}

\section{Related work}
\label{sec:related}

\textbf{Patch-based models.}
Inspired by the success of Transformers (or self-attention) \cite{vaswani2017attention} in natural language processing \cite{devlin2018bert,brown2020language}, numerous works have attempted to extend Transformers for computer vision \cite{khan2021transformers}. In particular, the seminal work named Vision Transformer (ViT) \cite{dosovitskiy2020image} discovered that Transformer could achieve state-of-the-art performance, alternating convolutional neural networks (CNNs) \cite{lecun1998gradient}. Other studies revealed that different patch mixing layers such as feed-forward \cite{tolstikhin2021mlp}, convolution \cite{trockman2022patches}, or even a pooling operation \cite{yu2021metaformer} are comparable to self-attention, hypothesizing that the success of ViT comes from the patch-based architectures. Our work proposes a universal framework to improve the patch-based models by introducing object-aware inductive bias to their patch mixing layers.

\textbf{Inductive bias for patch-based models.}
Many patch-based models aim to remove inductive biases by using patch mixing layers without additional structures, e.g., self-attention. While they perform well on large data regimes, recent works reveal that inductive biases are still crucial for patch-based models, especially under limited data \cite{steiner2021train}. Consequently, extensive literature proposed approaches to incorporate additional structures for patch-based models, e.g., spatial structures of CNNs. One line of work aims to design patch mixing layers reflecting inductive biases. For example, ConViT \cite{d2021convit} and CoAtNet \cite{dai2021coatnet} calibrate self-attention with spatial distances between patches, CvT \cite{wu2021cvt} and ConvMixer \cite{trockman2022patches} utilize convolution operation for patch mixing.
Another line of work builds an architecture that combines convolutional or pooling layers with patch mixing layers \cite{liu2021swin,yuan2021tokens,wang2021pyramid,fan2021multiscale,heo2021rethinking,yuan2021incorporating,xiao2021early}. Our work falls into the first category; however, we leverage the object structure of images, unlike prior works that focus on the spatial inductive bias. Using rich object information, our proposed OAMixer outperforms ConViT and CoAtNet, and using both gives further improvements, i.e., two methods contribute to the model differently (see Table~\ref{tab:bg}). We also emphasize that OAMixer can be applied to any patch mixing layers under a common principle, unlike prior works designed for specific layers. 

\textbf{Incorporating object structures.}
Although objects are the atom of scenes, only a limited number of research has leveraged the object structure of images for visual recognition.\footnote{
Some prior work \cite{dai2015convolutional,cheng2021masked} considered objects to modify features, but they simply masked the non-object regions, which loses information. We also tried a similar approach, but soft masking performed better.
} This is mainly due to two reasons: (a) the cost of object labels and (b) the non-triviality of reflecting object structure in neural networks. Recently, both challenges have been relaxed. First, the advance of unsupervised \cite{voynov2021object,caron2021emerging,mo2021object} and weakly-supervised \cite{selvaraju2017grad,chefer2021transformer,yun2021re} saliency detection significantly reduced the labeling cost of objects. Specifically, we use BigBiGAN \cite{voynov2021object} and ReLabel \cite{yun2021re} object labels in our experiments; no human annotations are required. Second, patch-based models are well-suited with object labels since the reduced resolution of patches (e.g., 16$\times$16 patches instead of 224$\times$224 pixels) enables the computation of their relations. OAMixer reflects the object structures into patch-based models in favor of these two advances. ORViT \cite{herzig2021object} directs video Transformers to focus on object regions, but they are less suited for image recognition. TokenLabeling \cite{jiang2021all} uses patch labels for auxiliary objectives; it can be combined with OAMixer and gives orthogonal gain (see Table~\ref{tab:bg}). We finally note that several works \cite{locatello2020object,wu2021generative,kipf2021conditional} aim to disentangle the object representation explicitly. However, they do not scale yet to complex real-world images due to the strong constraints in the model. In contrast, OAMixer can be applied to any existing patch-based models with minimal modification.

\section{OAMixer: Object-aware Mixing Layer}
\label{sec:method}

We first introduce OAMixer, an object-aware mixing layer for patch-based models in Section~\ref{subsec:method-OAMixer}. We then illustrate the instantiations of OAMixer on various mixing layers in Section~\ref{subsec:method-specific}. Finally, we discuss strategies to obtain patch-wise labels efficiently (e.g., unsupervised) in Section~\ref{subsec:method-label}.

\subsection{Computing reweighting mask for OAMixer}
\label{subsec:method-OAMixer}

The idea of patch-based models is to reshape a 2D image $\x \in \R^{H \times W \times C}$ into a sequence of flattened 2D patches $\x^0 \in \R^{N \times (P^2 C)}$, where $(H,W)$ is the resolution of original image, $C$ is the number of color channels, $(P,P)$ is the resolution of each image patch, and $N = HW/P^2$ is the number of patches. Patch-based models first convert the 2D patches into patch (or token) features $\x^1 := f_\texttt{embed}(\x^0) \in \R^{N \times D}$ with latent dimension $D$ using an embedding function $f_\texttt{embed}$, then update the patch features by alternating two operations: (a) patch mixing layers $f^l_\texttt{mix}: \R^N \to \R^N$ which mix the features among patches, and (b) channel mixing layers $g^l_\texttt{mix}: \R^D \to \R^D$ which mix the features among channels, where $l$ implies the operation of layer $l$. Formally, the $l$-th layer of patch-based model updates an input vector $\x^{l} \in \R^{N \times D}$ to an output vector $\x^{l+1} \in \R^{N \times D}$ following:
\begin{align}
\z^{l+1} = [\z^{l+1}_{1:N,d}] &= [f_\texttt{mix}(\x^l_{1,d} \,;\, \x^l_{2,d} \,;\, ... \,;\, \x^l_{N,d})] \\
\x^{l+1} = [\x^{l+1}_{n,1:D}] &= [g_\texttt{mix}(\z^{l+1}_{n, 1} \,;\, \z^{l+1}_{n, 2} \,;\, ... \,;\, \z^{l+1}_{n, D})]
\end{align}
where $\z_{n,d}$ and $\x_{n,d}$ denotes $n$-th patch, $d$-th channel value, and $\z^{l+1}_{1:N,d} \in \R^N$ and $\x^{l+1}_{n,1:D} \in \R^D$ denotes row-wise and column-wise subvector of $\z$ and $\x$, respectively.

We introduce OAMixer, a universal framework for improving patch mixing layers $f_\texttt{mix}$ by incorporating the object structure of images. With the object information, OAMixer strengthens the interaction of patches of similar objects while regularizing the connection of different objects and backgrounds. To this end, we utilize the patch-wise labels $\y \in \R^{N \times K}$ where $K$ is the number of object classes. Using them, we compute the reweighting mask $\M^l \in \R^{N \times N}$, a patch-wise semantical similarity mask.
Formally, we set the $(i,j)$-th value of the reweighting mask $\M^l_{ij}$ as a reverse distance between the object labels of two patches $\y_i$ and $\y_j$:
\begin{align}\label{eq:defM}
\M_{ij}^l := \exp(-\kappa^{(l)} \cdot d(\mathbf{y}_i, \mathbf{y}_j)) \in (0,1]
\end{align}
where $\kappa^{(l)} \ge 0$ is a learnable mask scale (scalar) parameter for each layer and $d(\cdot, \cdot): \R^K \times \R^K \to \R$ is a distance function for object labels. We initialize $\kappa^{(l)}$ by zero for training, i.e., consider the full interaction initially then focus on the objects as $\kappa^{(l)}$ increases. We observe that the model sets higher $\kappa^{(l)}$ for lower layers and lower $\kappa^{(l)}$ for higher layers after training, i.e., OAMixer automatically attends the intra-object relations first then expand to the inter-object relations, which resembles the structure of CNNs (see Table~\ref{tab:mask-scale}).

We calibrate the $N \times N$ interaction of patch mixing layers using the reweighting mask $\M$. If the patch mixing layer $f_\texttt{mix} := \LL_\texttt{mix}$ is linear, one can simply (element-wise) multiply the mask to get OAMixer; $\M \odot \LL_\texttt{mix}$. 
While OAMixer can be applied to any patch layers in principle, one needs careful design to consider the nonlinear dynamics of each layer. We provide specific implementations of OAMixer on various representative models in the next section.

\subsection{OAMixer for ViT and Mixers}
\label{subsec:method-specific}

We briefly review three representative patch mixing layers: self-attention, feed-forward, and convolution, and describe the implementations of OAMixer for each layer.

\textbf{OAMixer for self-attention.}
Self-attention \cite{vaswani2017attention} mixing layers update patch features by aggregating values with normalized importances (or attentions):
\begin{align}
f_\texttt{mix}(\x) := \underbrace{\texttt{Softmax}(\frac{\mathbf{Q} \mathbf{K}^T}{\sqrt{D_K}})}_\text{attention matrix $\mathbf{A}$} \cdot \mathbf{V}
\end{align}
where $\mathbf{Q}$, $\mathbf{K}$, $\mathbf{V}$ are query, key, and value, respectively, which are linear projections of input $\x \in \R^{N \times D}$, given by $\mathbf{Q} := \x \cdot \W_Q \in \R^{N \times D_K}$, $\mathbf{K} := \x \cdot \W_K \in \R^{N \times D_K}$, and $\mathbf{V} := \x \cdot \W_V \in \R^{N \times D_V}$. Here, we compute $H$ independent attention heads and aggregate outputs for the final output of size $N \times D$ for $D = H \cdot D_V$. Recall that self-attention is basically a matrix multiplication of attention $\A$ and value $\mathbf{V}$ matrices, and one can (element-wise) multiply the reweighting mask $\M$ to the attention matrix to calibrate interaction. Then, we renormalize the masked attention $\M \odot \A$ to make the row-wise sum be 1 as the original self-attention. To sum up, OAMixer for self-attention is:
\begin{align}
f_\texttt{oamix}(\x)
:= [\tilde{\A}_{ij}] \cdot \mathbf{V}
= [\frac{\M_{ij} \cdot \A_{ij}}{\sum_j \M_{ij} \cdot \A_{ij}}]\cdot \mathbf{V}
\label{eq:OAMixer-attn}
\end{align}
where $\tilde{\A}$ is the renormalized masked attention. We finally remark that patch-based models using self-attention often use the additional [CLS] token to aggregate the global feature. Here, we define the mask value between the [CLS] token and every other patch to be one and apply Eq.~\eqref{eq:OAMixer-attn}.

\textbf{OAMixer for feed-forward.}
Feed-forward (or multi-layer perceptron; MLP) mixing layers update patch features with a channel-wise MLP. Since each channel is computed independently, we only consider a $N \times 1$ vector of a single channel. Then, the mixer layer is:
\begin{align}
f_\texttt{mix}(\x) := \W_m \cdot \sigma(\W_{m-1} \cdot \sigma(\cdots \sigma(\W_1 \cdot \x)))
\end{align}
where $\W_1, ..., \W_m$ are weight matrices and $\sigma$ is a nonlinear activation. However, it is nontrivial to apply the reweighting mask $\M$ since $f_\texttt{mix}$ is nonlinear. To tackle this issue, we decompose the mixing layer $f_\texttt{mix}$ into a linear approximation $\LL_\texttt{mix}^\x \cdot \x \approx f_\texttt{mix}(\x)$ for a (possibly data-dependent) matrix $\LL_\texttt{mix}^\x \in \R^{N \times N}$ and a residual term $f_\texttt{mix}(\x) - \LL_\texttt{mix}^\x \cdot \x$. Here, we only calibrate the linear term but omit the residual term. Then, OAMixer for feed-forward is given by:
\begin{align}
f_\texttt{oamix}(\x) := \underbrace{(\M \odot \LL_\texttt{mix}^\x) \cdot \x}_\text{masked linear} + \underbrace{(f_\texttt{mix}(\x) - \LL_\texttt{mix}^\x \cdot \x)}_\text{residual}
\end{align}
where $\odot$ is an element-wise product. While finding a good $\LL_\texttt{mix}^\x$ is nontrivial in general, we found that simply dropping nonlinear activations gives an efficient yet effective solution:
\begin{align}
\LL_\texttt{mix} := \W_m ~ \W_{m-1} \cdots \W_1 \in \R^{N \times N}.
\end{align}
We observe that this (somewhat crude) approximation performs well. We also tried some data-dependent variants but did not see gain despite their computational burdens.

\textbf{OAMixer for convolution.}
Convolutional mixing layers update patch features with a channel-wise 2D convolution. Similar to the feed-forward case, we only consider a single channel input $\x$ (which is $\x_{1:N,d}$ formally), reshaped as a $1 \times \bar{H} \times \bar{W}$ tensor where $(\bar{H}, \bar{W}) = (H/P, W/P)$ is the resolution of patch features. Then, the mixer layer is:
\begin{align}
f_\texttt{mix}(\x) := \W_\texttt{conv} * \x
\end{align}
where $\W_\texttt{conv} \in \R^{1 \times 1 \times S \times S}$ is a kernel matrix with size $S$ and $*$ denotes convolution operation. Here, we consider the linearized version of kernel matrix (i.e., Toeplitz matrix) that substitutes the convolution to the matrix multiplication. Then, one can interpret the mixer layer as:
\begin{align}
f_\texttt{mix}(\x) = \W_\texttt{linear} \cdot \tilde{\x}
\end{align}
where $\W_\texttt{linear} \in \R^{N \times N}$ is the corresponding matrix of $\W_\texttt{conv}$ and $\tilde{\x}$ is a reshaped tensor of $\x$ of size $N \times 1$, where $N = \bar{H} \cdot \bar{W}$. Here, one can directly multiply the reweighting mask $\M$ to define the OAMixer for convolution:
\begin{align}
f_\texttt{oamix}(\x) := (\M \odot \W_\texttt{linear}) \cdot \tilde{\x}
\end{align}
where $\odot $ is an element-wise product. We also compare OAMixer with the models using different kernel matrix for each channel, i.e., $\W_\texttt{conv} \in \R^{D \times 1 \times S \times S}$ (see Appendix~\ref{appx:convmixer}).

\begin{table*}[t]
\caption{
Comparison of various scenarios evaluated on the Background Challenge benchmark. We report the classification accuracy (\buparrow) and background robustness (\rdownarrow), where higher and lower values are better for each metric. ``TL'' indicates TokenLabeling, that the model is trained with an additional token-level objective. Note that TL improves the overall performance; we colored them with gray lines for better visualization. We use ReLabel from ImageNet trained model for both OAMixer and TL. `+' denotes the modules added to the baseline (not accumulated), and parenthesis denotes the gap from the baseline. The table has several implications: (a) OAMixer consistently improves the classification and background robustness, confirming the benefit of object-centric representation. (b) OAMixer outperforms the methods using spatial inductive bias like ConViT and CoAtNet; combining ConViT and OAMixer performs even better. (c) OAMixer gives additional gain over TL, verifying the merit of guided patch interaction over na\"ive patch-level objective. (d) OAMixer also improves MLP-Mixer and ConvMixer, patch-based models using different patch mixing layers.
}\label{tab:bg}
\centering
\begin{tabular}{lclllll}
\toprule
& TL & Original (\buparrow) & Only-BG-B (\rdownarrow) & Mixed-Same (\buparrow) & Mixed-Rand (\buparrow) & BG-Gap (\rdownarrow)\\
\midrule
DeiT-T           & - & 70.62 & 35.48 & 63.31 & 39.56 & 23.75 \\
+ ConViT         & - & 78.74 \bup{8.12} & 38.69 \bup{3.21} & 70.84 \bup{7.53} & 46.91 \bup{7.35} & 23.93 \bup{0.18}  \\
+ CoAtNet        & - & 78.62 \bup{8.00} & 37.68 \bup{2.20} & 70.40 \bup{7.09} & 45.46 \bup{5.90} & 24.94 \bup{1.19}  \\
+ OAMixer (ours) & - & 81.98 \bup{11.36} & 30.89 \rdown{4.59} & 73.38 \bup{10.07} & 51.53 \bup{11.97} & 21.85 \rdown{1.90}  \\
+ OAMixer + ConViT (ours) & - & 83.38 \bup{12.76} & 31.36 \rdown{4.12} & 74.86 \bup{11.55} & 53.14 \bup{13.58} & 21.73 \rdown{2.02}  \\
\rowcolor{lightgray}
DeiT-T           & \checkmark & 83.50 & 37.01 & 74.57 & 51.78 & 22.79 \\
\rowcolor{lightgray}
+ OAMixer (ours) & \checkmark & 87.23 \bup{3.73} & 29.21 \rdown{7.80} & 78.15 \bup{3.58} & 60.59 \bup{8.81} & 17.56 \rdown{5.23} \\
\midrule
DeiT-S           & - & 82.69 & 39.46 & 73.88 & 50.49 & 23.39\\
+ ConViT         & - & 85.51 \bup{2.82} & 38.94 \rdown{0.52} & 76.40 \bup{2.52} & 54.37 \bup{3.88} & 22.03 \rdown{1.36}  \\
+ OAMixer (ours) & - & 86.32 \bup{3.63} & 31.78 \rdown{7.68} & 78.44 \bup{4.56} & 56.74 \bup{6.25} & 21.70 \rdown{1.69}  \\
+ OAMixer + ConViT (ours) & - & 88.20 \bup{5.51} & 30.79 \rdown{8.67} & 79.16 \bup{5.28} & 60.52 \bup{10.03} & 18.64 \rdown{4.75}  \\
\rowcolor{lightgray}
DeiT-S           & \checkmark & 88.52 & 38.00 & 79.43 & 59.36 & 20.07 \\
\rowcolor{lightgray}
+ OAMixer (ours) & \checkmark & 91.31 \bup{2.79} & 29.19 \rdown{8.81} & 81.78 \bup{2.35} & 66.92 \bup{7.56} & 14.86 \rdown{5.21} \\
\midrule
MLP-Mixer-S/16   & - & 84.99 & 40.96 & 76.52 & 54.72 & 21.80\\
+ OAMixer (ours) & - & 87.68 \bup{2.69} & 27.28 \rdown{13.68} & 79.43 \bup{2.91} & 60.44 \bup{5.72} & 18.99 \rdown{2.81}  \\
\rowcolor{lightgray}
MLP-Mixer-S/16   & \checkmark & 88.07 & 38.17 & 79.21 & 58.72 & 20.49 \\
\rowcolor{lightgray}
+ OAMixer (ours) & \checkmark & 90.49 \bup{2.42} & 28.72 \rdown{9.45} & 81.63 \bup{2.42} & 65.36 \bup{6.64} & 16.27 \rdown{4.22} \\
\midrule
ConvMixer-S/16   & - & 86.32 & 41.38 & 78.59 & 56.99 & 21.60\\
+ OAMixer (ours) & - & 88.49 \bup{2.17} & 35.60 \rdown{5.78} & 81.70 \bup{3.11} & 63.93 \bup{6.94} & 17.77 \rdown{3.83}  \\
\rowcolor{lightgray}
ConvMixer-S/16   & \checkmark & 87.49 & 36.49 & 80.42 & 58.15 & 22.27 \\
\rowcolor{lightgray}
+ OAMixer (ours) & \checkmark & 89.09 \bup{1.60} & 32.67 \rdown{3.82} & 82.64 \bup{2.22} & 63.70 \bup{5.55} & 18.94 \rdown{3.33}\\
\bottomrule
\end{tabular}
\end{table*}

\subsection{Obtaining object labels}
\label{subsec:method-label}

One possible concern on OAMixer is the annotation cost of object labels $\y \in \R^{N \times K}$. To alleviate this issue, we leverage the recent advances in unsupervised and weakly-supervised saliency detection. Specifically, we consider a two-stage training: train an object labeler on the downstream dataset, then train a OAMixer model (on the same dataset)
using the object labeler. Here, one can also use an object labeler pretrained under some external dataset like ImageNet \cite{deng2009imagenet}, instead of training a new object labeler for each downstream dataset.\footnote{The newly trained and ImageNet-pretrained object labelers show comparable performance in our experiments (see Table~\ref{tab:cls-more}).} In the remaining section, we recap the two types of machine annotators: binary saliency and multi-class prediction, and discuss their pros and cons.

\textbf{Binary saliency map.}
We first consider binary saliency maps, i.e., indicating whether the given pixel is an object or background. There is a tremendous amount of work on inferring saliency maps in unsupervised \cite{voynov2021object,caron2021emerging,mo2021object} or weakly-supervised (i.e., using class labels) \cite{selvaraju2017grad,chefer2021transformer} manners. We use an unsupervised method called BigBiGAN \cite{voynov2021object}, which finds the salient region using the latent space of BigGAN \cite{brock2019large}. We also use the attention map of pretrained DINO \cite{caron2021emerging} model to extract the object information unsupervised. From the binary saliency map, we get a soft label $\y_n \in [0,1]$, and use the $l_1$-distance between the object labels of two patches in Eq.~\eqref{eq:defM}.

\textbf{Multi-class prediction map.}
We also consider multi-class prediction maps, i.e., pixel- or patch-level semantic masks. In particular, we use ReLabel \cite{yun2021re}, a weakly-supervised method that predicts dense label maps from an image classifier by applying the classifier to the penultimate spatial features (i.e., before global average pooling). We then resize the label map (by pooling over interpolated values) to match the resolution of tokens and get the object labels $\y_n \in \mathbb{R}^K$. Here, we use the cosine distance to compute the distance between object labels of two patches.

\textbf{Comparison of two approaches.}
Multi-class prediction map contains richer semantics and thus provides more informative reweighing masks. As a result, OAMixer using ReLabel performs better than BigBiGAN in most cases. In particular, we use multi-class prediction maps for multi-object images since binary saliency maps cannot distinguish different objects. However, BigBiGAN works better on distribution shifts, as finding salient regions is easier than predicting semantics for the unseen images. Thus, the users can choose binary or multi-class labels on their purpose.

\begin{table*}[t]
\caption{
ImageNet top-1 accuracy. We add TokenLabeling (TL) and OAMixer on top of the baseline models (accumulated). Both TL and OAMixer use ReLabel object labels from ImageNet-trained model. Bold denotes the best result, and parenthesis denotes the gap from the baseline. Note that TL requires enough model capacity (better for the larger model, even harms MobileViT-XXS). In contrast, OAMixer gives a consistent improvement over all considered model capacities.
}\label{tab:cls-imagenet}
\centering
\begin{tabular}{lcccc}
\toprule
& MobileViT-XXS & DeiT-T & DeiT-S & DeiT-B \\
\midrule
\# params & 1.3M & 5M & 22M & 86M \\
\midrule
Baseline & 62.56 \blank & 71.19 \blank & 79.30 \blank & 78.45 \blank \\
+ TokenLabeling & 62.34 \down{0.22} & 71.61 \up{0.42} & 80.20 \up{0.90} & 81.17 \up{2.72} \\
+ OAMixer (ours) & \textbf{64.78} \up{2.22} & \textbf{74.26} \up{3.07} & \textbf{81.26} \up{1.96} & \textbf{82.18} \up{3.73} \\
\bottomrule
\end{tabular}
\end{table*}

\begin{table*}[t]
\caption{
Linear probing results of DINO with DeiT-T trained on ImageNet with 100 epochs. We use the unsupervised attention map from the pretrained baseline DINO model as a patch-wise object label for OAMixer. We test the in-domain accuracy with the ImageNet validation dataset and out-domain accuracy with Background Challenge datasets after training the linear classifier on the ImageNet train set with 100 epochs. The results show that OAMixer is effective in the self-supervised framework.
}\label{tab:dino}
\centering
\begin{tabular}{llllll}
\toprule
& ImageNet (\buparrow) & Only-BG-B (\rdownarrow) & Mixed-Same (\buparrow) & Mixed-Rand (\buparrow) & BG-Gap (\rdownarrow)\\
\midrule
DeiT-T            & 59.37 & 34.27 & 89.73 & 78.07 & 11.66 \\
+ OAMixer          & 61.16 \bup{1.79} & 20.20 \rdown{14.07} & 89.75 \bup{0.02} & 79.33 \bup{1.26} & 10.42 \rdown{1.24} \\
\bottomrule
\end{tabular}
\end{table*}

\section{Experiments}
\label{sec:exp}
We first show that the object-centric design (i.e., disentagled representation of objects and background) of OAMixer improves the classification accuracy and background robustness of various patch-based models in Section~\ref{subsec:exp-main}. We then verify the applicability of OAMixer over various downstream tasks, including large-scale classification, self-supervised learning, and multi-object recognition in Section~\ref{subsec:exp-compare}. We finally provide analyses in Section~\ref{subsec:exp-analysis}.

\textbf{Models and training.}
We apply OAMixer on three representative patch-based models: DeiT \cite{touvron2021training}, MLP-Mixer \cite{tolstikhin2021mlp}, and ConvMixer \cite{trockman2022patches}, using self-attention, feed-forward, and convolutional patch mixing layers, respectively. Specifically, we use DeiT-T, DeiT-S, MLP-Mixer-S/32, and MLP-Mixer-S/16 configurations. For ConvMixer, we follow the configuration of MLP-Mixer (dim: 512, depth: 8) and use kernel size of 9 for a fair comparison with MLP-Mixer. We denote these variants as ConvMixer-S/32 and ConvMixer-S/16. We share the convolution kernel over channels by default, while we provide the unshared results in Appendix~\ref{appx:convmixer}. We follow the default training setup of DeiT, but use a batch size of 256 due to GPU memory issue.

\textbf{Object labels.}
We consider binary and multi-class prediction maps, as explained in Section~\ref{subsec:method-label}. The object labelers are either trained on each downstream dataset or pretrained on ImageNet \cite{deng2009imagenet}. We train a ResNet-18 \cite{he2016deep} network to obtain ReLabel on the downstream dataset and use the BigBiGAN and ReLabel with NFNet-F6 \cite{brock2021high} trained on the ImageNet dataset for pretrained object labelers.

\begin{table*}[t]
\caption{
Classification accuracy on various distribution-shifted datasets. Bold denotes the best results, and parenthesis denotes the gap from the baseline. OAMixer gives consistent improvements.
}\label{tab:dist-shift}
\centering
\begin{tabular}{ll|lllll}
\toprule
& ImageNet-9 & ImageNetV2-9 & ReaL-9 & Rendition-9 & Stylized-9 & Sketch-9 \\
\midrule
DeiT-T           & 70.62 & 63.19 & 68.13 & 24.14 & 16.66 & 21.65 \\
+ OAMixer (ours) & \textbf{80.15} \up{9.53} & \textbf{70.68} \up{7.49} & \textbf{77.78} \up{9.65} & \textbf{27.53} \up{3.39} & \textbf{21.80} \up{5.14} & \textbf{24.45} \up{2.80} \\
\midrule
DeiT-S           & 82.69 & 73.22 & 80.21 & 28.60 & 21.90 & 27.26 \\
+ OAMixer (ours) & \textbf{83.88} \up{1.19} & \textbf{75.68} \up{2.46} & \textbf{82.37} \up{2.16} & \textbf{29.43} \up{0.83} & \textbf{24.64} \up{2.74} & \textbf{27.93} \up{0.67} \\
\bottomrule
\end{tabular}
\end{table*}

\begin{table*}[t]
\caption{
Classification accuracy on various datasets. We add OAMixer to the baseline models, where ``OAMixer'' uses ReLabel trained on each downstream dataset, and ``OAMixer$^*$'' uses ReLabel trained on ImageNet. Bold denotes the best result among methods not using ImageNet data (not colored by gray lines), and parenthesis denotes the gap from the baseline. OAMixer gives a consistent improvement, showing comparable results with OAMixer$^*$ for all considered datasets.
}\label{tab:cls-more}
\centering
\begin{tabular}{lccgccg}
\toprule
& \multicolumn{3}{c}{DeiT-T} & \multicolumn{3}{c}{DeiT-S} \\
\cmidrule(lr){2-4}\cmidrule(lr){5-7} 
& Baseline & OAMixer (ours) & OAMixer$^*$ (ours) & Baseline & OAMixer (ours) & OAMixer$^*$ (ours) \\
\midrule
CIFAR-10  & 86.35 & \textbf{89.89} \up{3.54} & 88.09 \up{1.74} & 90.80 & \textbf{92.73} \up{1.93} & 92.33 \up{1.53} \\
CIFAR-100 & 65.21 & \textbf{66.94} \up{1.73} & 67.21 \up{2.00} & 70.08 & \textbf{71.00} \up{0.92} & 71.63 \up{1.55} \\
Food-101  & 69.83 & \textbf{72.25} \up{2.42} & 71.50 \up{1.67} & 75.00 & \textbf{75.17} \up{0.17} & 76.03 \up{1.03} \\
Pets & 24.20 & \textbf{24.50} \up{0.30} & 24.97 \up{0.57} & 25.57 & \textbf{30.64} \up{5.07} & 29.63 \up{4.06} \\
\bottomrule
\end{tabular}
\end{table*}


\subsection{Main results}
\label{subsec:exp-main}
In this section, we show that the object-aware inductive bias from OAMixer improves classification and background robustness over various patch-based models. To this end, we conduct experiments on the Background Challenge \cite{xiao2020noise} benchmark, which consists of a smaller (9 superclass, 370 class) subset of ImageNet as a training dataset, called ImageNet-9 (IN-9), and eight test datasets to evaluate the classification and background robustness.\footnote{
We also provide the results using a larger version of IN-9 (IN-9L) as a training dataset in Appendix~\ref{appx:in9l}. It improves the overall performance of all models but shows a similar tendency with IN-9.
} In this section, we will demonstrate that (a) OAMixer improves classification and background robustness over various patch-based models, OAMixer outperforms (b) prior methods using spatial inductive bias, and (c) prior methods using patch-level supervision. In addition, we compare OAMixer using different object labels including BigBiGAN in Appendix~\ref{appx:bigbigan}.

\textbf{Setup.}
Background Challenge contains eight evaluation datasets with different combinations of foregrounds and backgrounds: \textsc{Original}~($\uparrow$), \textsc{Only-BG-B}~($\downarrow$), \textsc{Only-BG-T}~($\downarrow$), \textsc{No-FG}~($\downarrow$), \textsc{Only-FG}~($\uparrow$),  \textsc{Mixed-Same}~($\uparrow$), \textsc{Mixed-Rand}~($\uparrow$), \textsc{Mixed-Next}~($\uparrow$), where the upper or lower arrows imply the model should predict the class well or not, respectively. Namely, the model should increase the accuracy of ($\uparrow$) datasets while decrease the accuracy (or increase the background robustness) of ($\downarrow$) datasets. We also report \textsc{BG-Gap}~($\downarrow$) which measures the accuracy gap between \textsc{Mixed-Same} and \textsc{Mixed-Rand}. We omit \textsc{Only-BG-T}, \textsc{No-FG}, \textsc{Only-FG}, and \textsc{Mixed-Next} results for the brevity (see Appendix~\ref{appx:bg_challenge} for discussion). We use ReLabel trained on ImageNet in the experiments, but ReLabel trained on IN-9 gives similar trends (see Appendix~\ref{appx:relabel_in9}).

\textbf{Classification and background robustness.}
Table~\ref{tab:bg} presents the results on the Background Challenge benchmark, which has several additional implications that will be discussed in a few following paragraphs. First, we remark that OAMixer consistently improves the classification and background robustness in all considered patch-based models. To highlight a few numbers, OAMixer enhances the classification accuracy of DeiT-T from 70.62\% to 81.98\% (+11.36\%) on the \textsc{Original} dataset while reducing the background bias of MLP-Mixer-S/16 from  40.96\% to 27.28\% (-13.68\%) on the \textsc{Only-BG-B} dataset. These results support that OAMixer learns more object-centric representation robust to background shifts.

\textbf{Comparison with methods using spatial inductive bias.}
We compare OAMixer with patch mixing layers using spatial inductive bias: ConViT \cite{d2021convit} and CoAtNet \cite{dai2021coatnet}. We only take the modified self-attention layers from both models and apply them to DeiT for a fair comparison with OAMixer. ConViT and CoAtNet calibrate the self-attention but use spatial distance instead of semantic distance from object labels. We keep the last two layers of ConViT as the vanilla self-attention following the original paper. Table~\ref{tab:bg} shows that OAMixer outperforms both ConViT and CoAtNet for all considered scenarios. Notably, the DeiT-T results show that spatial inductive bias often does not help background robustness, as local aggregation does not prevent the entangling of patches in adjacent regions. We also remark that combining ConViT and OAMixer performs the best; they give orthogonal benefits.

\textbf{Comparison with methods using patch-level supervision.}
We compare OAMixer with prior work using patch-level supervision, TokenLabeling (TL) \cite{jiang2021all}. Specifically, TL uses the object (or patch) labels as an additional objective for each patch. We remark that TL and OAMixer benefit the model differently, and using both performs the best. Table~\ref{tab:bg} shows that OAMixer+TL outperforms TL for all considered scenarios. It confirms that the advantage of OAMixer comes from the guided patch interaction, not only from the usage of patch-level supervision. In particular, we highlight that TL does not improve the background robustness, unlike OAMixer gives a significant gain.


\subsection{Various downstream tasks}
\label{subsec:exp-compare}

In this section, we vefity the benefits of OAMixer on various computer vision tasks, including ImageNet classification, self-supervised learning, and multi-object recognition. To do so, we conduct experiments on the ViT and apply OAMixer on self-attention since ViTs are the most widely used patch-based models in various applications.

\textbf{ImageNet classification.}
Table~\ref{tab:cls-imagenet} presents the ImageNet \cite{deng2009imagenet} top-1 accuracy of OAMixer and baseline models on various ViT architectures: MobileViT and DeiT. For a fair comparison, we also compare OAMixer with TokenLabeling (TL) \cite{jiang2021all}, which also uses object labels for training. We use ReLabel from ImageNet trained model for both methods. The results show that OAMixer+TL outperforms TL for all considered scenarios. Note that the gain of TL is proportional to the model capacity, i.e., more beneficial for the larger model and even harmful for the small model, MobileViT-XXS. We think this is because the additional supervision of TL puts more burden on the models, requiring a larger capacity to digest. In contrast, OAMixer modifies the optimization landscape via reparametrization of patch mixing layers, leading the models to converge to the better optima regardless of the model capacity. Also, note that OAMixer regularizes the overfitting of the vanilla DeiT-B.

\textbf{Self-supervised learning.} Table~\ref{tab:dino} shows that OAMixer is also effective for self-supervised learning. Here, we apply OAMixer on top of the DINO \cite{caron2021emerging} framework, which are known to effective for learning representations for ViT.
To this end, we first train the vanilla DINO with DeiT-T on ImageNet. We then train DINO with OAMixer, using the attention map of the pretrained vanilla baseline DINO model as unsupervised binary patch-wise object labels. To evaluate the representation, we train a linear classifier for 9 classes of Background Challenge upon the learned representation. DINO trained with OAMixer learns more object-centric representation, and highly improves the classification accuracy and background robustness.

\textbf{Generalization on distribution shifts.} Table~\ref{tab:dist-shift} shows the classification accuracy of models trained on ImageNet-9, evaluated on various distribution-shifted datasets: 9 superclass (370 class) subsets of ImageNetV2~\cite{recht2019imagenet}, ImageNet-ReaL~\cite{beyer2020we}, ImageNet-Rendition~\cite{hendrycks2021many}, ImageNet-Stylized~\cite{geirhos2018imagenet}, and ImageNet-Sketch~\cite{wang2019learning}, denoted by adding `-9' at the suffix of their names. OAMixer also generalizes well on distribution shifts. We use BigBiGAN for object labels since it worked better than ReLabel; finding salient regions is easier than predicting semantics for the unseen samples.

\textbf{More classification results.}
Table~\ref{tab:cls-more} presents the classification accuracy on various downstream datasets, including CIFAR-\{10,100\}~\cite{krizhevsky2009learning}, Food-101~\cite{bossard2014food}, and Pets~\cite{parkhi2012cats}. We modified ViT architectures for CIFAR to use 2$\times$2 patch size following Trockman et al. \cite{trockman2022patches}, as CIFAR has a smaller resolution of 32$\times$32 than the original architecture for 224$\times$224. Here, we compare two versions of OAMixer: ``OAMixer'' using ReLabel trained on each downstream dataset, and ``OAMixer$^*$'' using ReLabel trained on ImageNet. We observe that ReLabels trained on the downstream task and ImageNet show comparable performance. We think this is because of the distribution shift on the downstream task, which compensates for the extra data of ImageNet. Thus, one can simply use the ImageNet-pretrained ReLabel instead of training a specific object labeler for each dataset.


\begin{table}[t]
\caption{
Mean average precision (mAP) for multi-object recognition on the COCO dataset. We use the patch-wise object labels, which are given by ReLabel trained on ImageNet. Bold denotes the best results, and parenthesis denotes the gap from the baseline. OAMixer is also effective for recognizing multi-object images.
}\label{tab:cls-multi}
\centering\small
\begin{tabular}{cccc}
\toprule
\multicolumn{2}{c}{DeiT-T} & \multicolumn{2}{c}{DeiT-S} \\
\cmidrule(lr){1-2}\cmidrule(lr){3-4} 
Baseline & OAMixer (ours) & Baseline & OAMixer (ours) \\
\midrule
57.42 & \textbf{59.56} \up{2.14} & 60.93 & \textbf{62.23} \up{1.30} \\
\bottomrule
\end{tabular}
\end{table}

\textbf{Multi-object recognition.}
Table~\ref{tab:cls-multi} presents the mean average precision (mAP) results on the COCO multi-object recognition task, which aims to predict the occurrence of each class for multi-object images. Here, we use multi-class prediction maps for OAMixer since binary saliency maps cannot distinguish different objects. Specifically, we use ReLabel from ImageNet trained model. The results confirm that OAMixer is also effective for multi-object images.


\subsection{Analysis}
\label{subsec:exp-analysis}
\begin{table}[t]
\captionof{table}{
Learned mask scales $\kappa^{(l)}$ over layers.
}\label{tab:mask-scale}
\centering
\resizebox{.95\linewidth}{!}{%
\begin{tabular}{lcccc}
\toprule
& Layer 1/4 & Layer 2/4 & Layer 3/4 & Layer 4/4 \\
\midrule
DeiT-S         & 1.571 & 0.783 & 0.945 & 0.287 \\
MLP-Mixer-S/16 & 0.744 & 0.000 & 0.001 & 0.061 \\
ConvMixer-S/16 & 0.871 & 2.347 & 1.498 & 0.001 \\
\bottomrule
\end{tabular}}
\end{table}

\textbf{Mask scales.}
We report the mask scales $\kappa^{(l)}$ in Eq.~\eqref{eq:defM} of trained models in Table~\ref{tab:mask-scale}. We divide the model quarterly by depth and average the values for each quarter. Remark that the models set higher $\kappa^{(l)}$ for lower layers and lower $\kappa^{(l)}$ for higher layers (especially $\kappa^{(l)}=0$ for the final layer), i.e., see the objects first then expand its view, resembling the local-to-global structure of CNNs. ConvMixer sets low $\kappa^{(l)}$ for the early layers since it is hard to understand the objects due to the restricted view of convolution.

\begin{figure}[t]
\centering
\begin{subfigure}{0.13\textwidth}
\includegraphics[width=\textwidth]{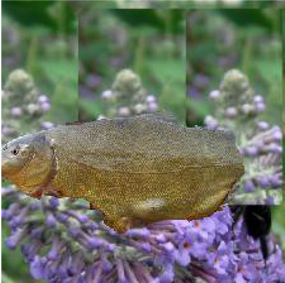}
\caption{Original}
\end{subfigure}
\begin{subfigure}{0.13\textwidth}
\includegraphics[width=\textwidth]{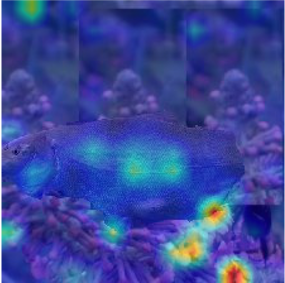}
\caption{DeiT-S}
\end{subfigure}
\begin{subfigure}{0.13\textwidth}
\includegraphics[width=\textwidth]{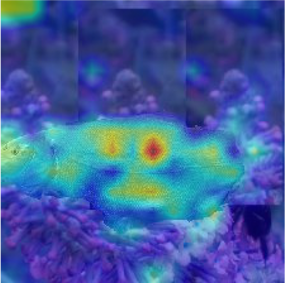}
\caption{OAMixer}
\end{subfigure}
\caption{
Saliency maps of DeiT-S and OAMixer (ours).
}\label{fig:saliency}
\end{figure}

\textbf{Saliency maps.}
We visualize the saliency maps \cite{chefer2021transformer} of DeiT-S and OAMixer on top of it, trained on ImageNet-9. Figure~\ref{fig:saliency} shows the saliency maps on a background-shifted image from the \textsc{Mixed-Rand} dataset, indicating that OAMixer sees a more object-centric view. Quantitatively, OAMixer improves the mean intersection-over-union (mIoU) of saliency maps of DeiT-S by 10\% relatively.

\section{Conclusion}
\label{sec:conclusion}

We propose OAMixer, a novel object-centric framework to refine patch-based models. We demonstrate that OAMixer outperforms the classification and background robustness of various patch-based models. We also show the effectiveness of OAMixer on various downstream tasks. We hope OAMixer could inspire new research directions for patch-based models and object-centric learning.

\section*{Acknowledgement}

This research was supported by the Engineering Research Center Program through the National Research Foundation of Korea (NRF) funded by the Korean Government MSIT (NRF2018R1A5A1059921).

{\small
\bibliographystyle{ieee_fullname}
\bibliography{cvprw_ref}
}

\clearpage
\appendix

\onecolumn

\appendix
\section{Discussions on Background Challenge}
\label{appx:bg_challenge}

\begin{figure*}[ht]
\centering
\includegraphics[width=.8\textwidth]{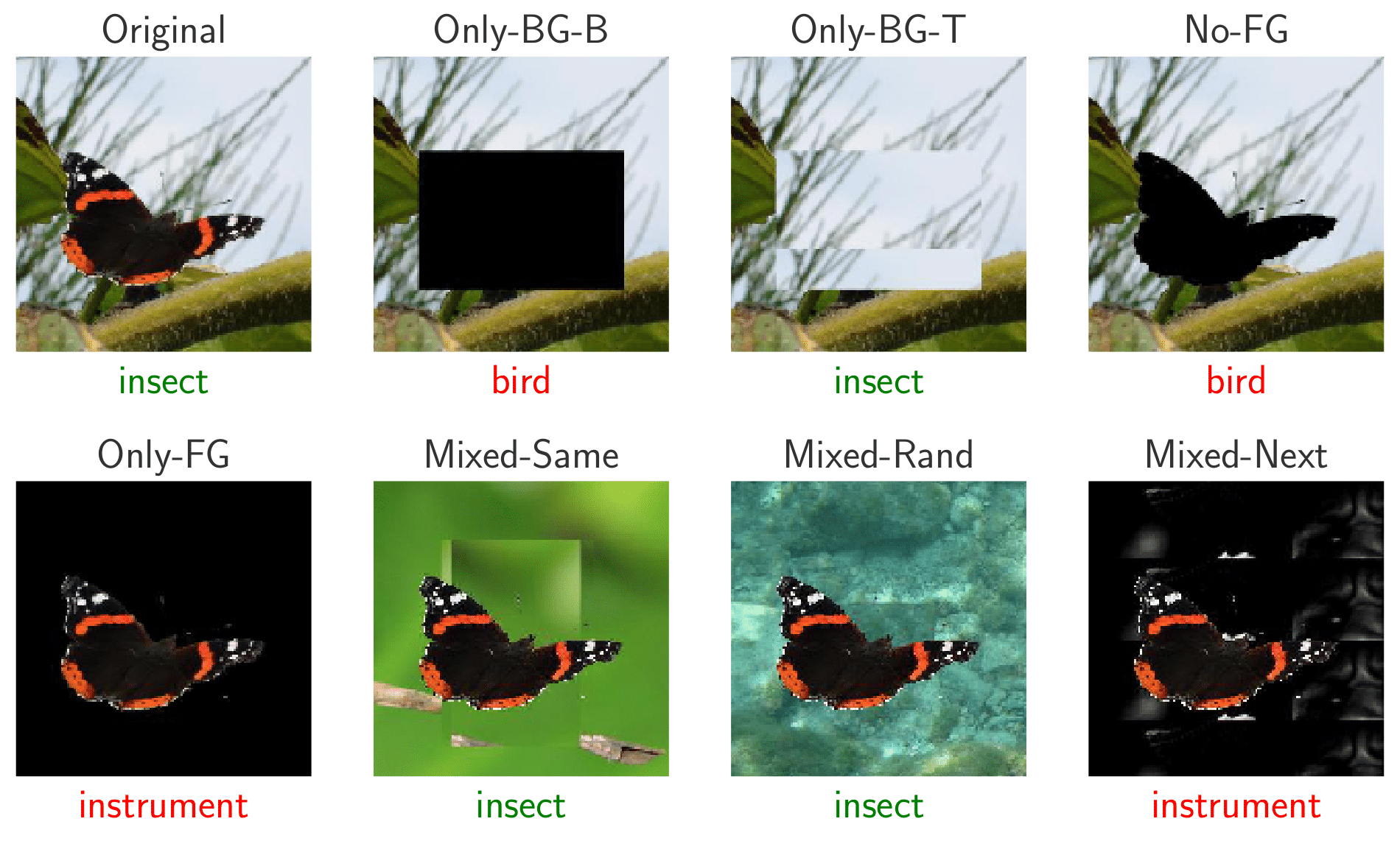}
\caption{
Datasets in the Background Challenge benchmark.
}\label{fig:bg_challenge}
\end{figure*}

Figure~\ref{fig:bg_challenge} visualizes the datasets in the Background Challenge \cite{xiao2020noise} benchmark, which provides various combinations of foregrounds and backgrounds: \textsc{Original}~($\uparrow$), \textsc{Only-BG-B}~($\downarrow$), \textsc{Only-BG-T}~($\downarrow$), \textsc{No-FG}~($\downarrow$), \textsc{Only-FG}~($\uparrow$), \textsc{Mixed-Same}~($\uparrow$), \textsc{Mixed-Rand}~($\uparrow$), \textsc{Mixed-Next}~($\uparrow$). The upper or lower arrows indicate whether the model should predict the class well or not, respectively. We omit \textsc{Only-BG-T},  \textsc{No-FG}, \textsc{Only-FG}, and \textsc{Mixed-Next} results due to the brevity of the presentation. Nevertheless, we provide some observations and discussions on them:
\begin{itemize}[topsep=1.0pt,itemsep=1.0pt,leftmargin=5.5mm]
    \item[$\circ$] \textsc{Only-FG} and \textsc{Mixed-Next} shows the same trend of \textsc{Mixed-Rand}.
    \item[$\circ$] \textsc{No-FG} is controversial to be predicted or not since the image contains the shadow of the object.
    \item[$\circ$] \textsc{Only-BG-T} increases when \textsc{Original} increases. However, OAMixer only increases \textsc{Only-BG-T} a little while \textsc{Original} a lot compared to the baselines.
\end{itemize}

\clearpage
\section{OAMixer using different object labels}
\label{appx:bigbigan}

\begin{table*}[ht]
\caption{
Comparison of OAMixer using various object labels. We test three different object labels: (a) BigBiGAN \cite{voynov2021object} (unsupervised binary saliency map), (b) ReLabel \cite{yun2021re} (weakly-supervised multi-class map) and (c) Bbox + Grabcut \cite{rother2004grabcut} (binary saliency map extracted from the ground-truth bounding box). We report the classification accuracy (\buparrow) and background robustness (\rdownarrow), where higher and lower values are better for each metric. `+' denotes the modules added to the baseline (not accumulated), and parenthesis denotes the gap from the baseline. The table shows OAMixer consistently outperforms the baseline for all considering object labelers.
}\label{tab:bigbigan}
\centering
\begin{tabular}{lllllll}
\toprule
& Object label & Original (\buparrow) & Only-BG-B (\rdownarrow) & Mixed-Same (\buparrow) & Mixed-Rand (\buparrow) & BG-Gap (\rdownarrow)\\
\midrule
DeiT-T           & -            & 70.62 & 35.48 & 63.31 & 39.56 & 23.75 \\
+ OAMixer (ours) & BigBiGAN     & 80.15 \bup{9.53} & 32.37 \rdown{3.11} & 71.98 \bup{8.66} & 48.77 \bup{9.21} & 23.21 \rdown{0.54}  \\
+ OAMixer (ours) & ReLabel      & 81.98 \bup{11.36} & 30.89 \rdown{4.59} & 73.38 \bup{10.07} & 51.53 \bup{11.97} & 21.85 \rdown{1.90}  \\
+ OAMixer (ours) & Bbox+GrabCut & 83.23 \bup{12.61} & 34.49 \rdown{0.99} & 74.32 \bup{11.01} & 52.77 \bup{13.21} & 21.55 \rdown{2.20}  \\
\midrule
DeiT-S           & -            & 82.69 & 39.46 & 73.88 & 50.49 & 23.39\\
+ OAMixer (ours) & BigBiGAN     & 83.88 \bup{1.19} & 36.59 \rdown{2.87} & 75.95 \bup{2.07} & 54.07 \bup{3.58} & 21.88 \rdown{1.51}  \\
+ OAMixer (ours) & ReLabel      & 86.32 \bup{3.63} & 31.78 \rdown{7.68} & 78.44 \bup{4.56} & 56.74 \bup{6.25} & 21.70 \rdown{1.69}  \\
+ OAMixer (ours) & Bbox+GrabCut & 87.04 \bup{4.35} & 37.80 \rdown{1.66} & 78.15 \bup{4.27} & 57.80 \bup{7.31} & 20.35 \rdown{3.04}  \\
\midrule
MLP-Mixer-S/16   & -            & 84.99 & 40.96 & 76.52 & 54.72 & 21.80\\
+ OAMixer (ours) & BigBiGAN     & 86.30 \bup{1.31} & 36.64 \rdown{4.32} & 78.47 \bup{1.95} & 58.54 \bup{3.82} & 19.93 \rdown{1.87}  \\
+ OAMixer (ours) & ReLabel      & 87.68 \bup{2.69} & 27.28 \rdown{13.68} & 79.43 \bup{2.91} & 60.44 \bup{5.72} & 18.99 \rdown{2.81}  \\
+ OAMixer (ours) & Bbox+GrabCut & 88.32 \bup{3.33} & 37.51 \rdown{3.45} & 79.68 \bup{3.16} & 61.85 \bup{7.13} & 17.83 \rdown{3.97}  \\
\midrule
ConvMixer-S/16   & -            & 86.32 & 41.38 & 78.59 & 56.99 & 21.60\\
+ OAMixer (ours) & BigBiGAN     & 86.35 \bup{0.03} & 37.09 \rdown{4.29} & 80.27 \bup{1.68} & 59.19 \bup{2.20} & 21.09 \rdown{0.51}  \\
+ OAMixer (ours) & ReLabel      & 88.49 \bup{2.17} & 35.60 \rdown{5.78} & 81.70 \bup{3.11} & 63.93 \bup{6.94} & 17.77 \rdown{3.83}  \\
+ OAMixer (ours) & Bbox+GrabCut & 87.73 \bup{1.41} & 34.64 \rdown{6.74} & 80.03 \bup{1.44} & 61.88 \bup{4.89} & 18.15 \rdown{3.45}  \\
\bottomrule
\end{tabular}
\end{table*}

Table~\ref{tab:bigbigan} compares OAMixer using BigBiGAN and ReLabel. We also compare with Bbox+Grabcut, a binary saliency map extracted from the bounding boxes using the GrabCut \cite{rother2004grabcut} algorithm, which is provided by the Background Challenge benchmark. Bbox+GrabCut gives accurate object masks as it leverages the bounding boxes. Here, one can see the differences between object labels. First, Bbox+GrabCut performs the best for most cases, supporting that the accurate object labels are helpful for OAMixer. Second, ReLabel performs the best for the \textsc{Only-BG-B} dataset; this is because the binary saliency models predict the background-only images as all-zero values, reducing OAMixer behaves like the vanilla patch-based model. We use ReLabel as a default choice in our experiments, as it does not require additional human annotations yet performs well.

\section{ReMixer using ReLabel from ImageNet-9}
\label{appx:relabel_in9}

\begin{table*}[h]
\caption{
Background Challenge results of ReMixer using ReLabel trained on ImageNet-9 (IN-9) and ImageNet (IN-1K). ReLabel trained on IN-9 shows similar trend with IN-1K.
}\label{tab:relabel_in9}
\vspace{-0.1in}
\centering\small
\begin{tabular}{llllllll}
\toprule
& ReLabel & Original (\buparrow) & Only-BG-B (\rdownarrow) & Mixed-Same (\buparrow) & Mixed-Rand (\buparrow) & BG-Gap (\rdownarrow)\\
\midrule
DeiT-T           & & 70.62 & 35.48 & 63.31 & 39.56 & 23.75 \\
+ ReMixer (ours) & IN-9  & 79.26 \bup{8.64} & 33.63 \rdown{1.85} & 70.94 \bup{7.63} & 47.38 \bup{7.82} & 23.56 \rdown{0.19} \\
+ ReMixer (ours) & IN-1K & 81.98 \bup{11.36} & 30.89 \rdown{4.59} & 73.38 \bup{10.07} & 51.53 \bup{11.97} & 21.85 \rdown{1.90}  \\
\bottomrule
\end{tabular}
\end{table*}

Table~\ref{tab:relabel_in9} compares ReMixer using ReLabel trained on ImageNet-9 (IN-9) and ImageNet (IN-1K). ReLabel trained on IN-1K performs better but shows the same trends with IN-9.

\clearpage
\section{ImageNet-9 Large (IN-9L) results}
\label{appx:in9l}

\begin{table*}[h]
\caption{
Background Challenge results of models trained on IN-9 and IN-9L datasets. While training on IN-9L significantly improves the performance overall, it shows a consistent trend with IN-9.
}\label{tab:in9l}
\vspace{-0.1in}
\centering\small
\begin{tabular}{lllllll}
\toprule
& Original (\buparrow) & Only-BG-B (\rdownarrow) & Mixed-Same (\buparrow) & Mixed-Rand (\buparrow) & BG-Gap (\rdownarrow)\\
\midrule
\multicolumn{6}{c}{\cellcolor{lightgray} \textit{Trained on IN-9}} \\
\midrule
DeiT-T           & 70.62 & 35.48 & 47.04 & 39.56 & 23.75 \\
+ ReMixer (ours) & 81.98 \bup{11.36} & 30.89 \rdown{4.59} & 54.03 \bup{6.99} & 51.53 \bup{11.97} & 21.85 \rdown{1.90}  \\
\midrule
\multicolumn{6}{c}{\cellcolor{lightgray} \textit{Trained on IN-9L}} \\
\midrule
DeiT-T           & 93.38 & 41.93 & 77.75 & 68.15 & 17.21 \\
+ ReMixer (ours) & 95.31 \bup{1.93} & 32.32 \rdown{9.61} & 81.16 \bup{3.41} & 74.44 \bup{6.29} & 14.00 \rdown{3.21} \\
\bottomrule
\end{tabular}
\end{table*}

Table~\ref{tab:in9l} compares the models trained on ImageNet-9 (IN-9) and the larger version of IN-9 (IN-9L). The models trained on IN-9L performs better but shows the same trends with IN-9.

\section{Sharing weights for ConvMixer}
\label{appx:convmixer}

\begin{table*}[ht]
\caption{
Comparison of ConvMixer sharing (white line) and not sharing (gray line) kernels over channels. We add `-Full' to denote the latter, which uses $\times D$ parameters for patch mixing layers where $D$ is the number of channels. ConvMixer + OAMixer is comparable with ConvMixer-Full.
}\label{tab:convmixer}
\centering
\begin{tabular}{llllllll}
\toprule
& Original (\buparrow) & Only-BG-B (\rdownarrow) & Mixed-Same (\buparrow) & Mixed-Rand (\buparrow) & BG-Gap (\rdownarrow)\\
\midrule
ConvMixer-S/16   & 86.32 & 41.38 & 66.84 & 56.99 & 21.60\\
+ OAMixer (ours) & 88.49 \bup{2.17} & \textbf{35.60} \rdown{5.78} & \textbf{71.60} \bup{4.76} & \textbf{63.93} \bup{6.94} & \textbf{17.77} \rdown{3.83}  \\
ConvMixer-S/16-Full & \textbf{88.67} \bup{2.35} & 40.20 \rdown{1.18} & 70.72 \bup{3.88} & 62.84 \bup{5.85} & 18.47 \rdown{3.13}  \\
\bottomrule
\end{tabular}
\end{table*}

We adopt ConvMixer \cite{trockman2022patches} to share the convolution kernels over channels to reduce memory and computation in our experiments. Table~\ref{tab:convmixer} shows the comparison of our adopted ConvMixer and the original one (ConvMixer-Full). ConvMixer + OAMixer is comparable or even outperforms ConvMixer-Full, yet using $D=512$ times smaller parameters for patch mixing layers.

\end{document}